\documentclass[10pt,twocolumn,letterpaper]{article}

\usepackage{cvpr}              % To produce the CAMERA-READY version
\definecolor{cvprblue}{rgb}{0.21,0.49,0.74}
\usepackage[pagebackref,breaklinks,colorlinks,allcolors=cvprblue]{hyperref}
\usepackage{comment}
\usepackage{amssymb}  % For \checkmark
\usepackage{pifont}
\usepackage{marvosym} % For \xmark
\usepackage{graphicx}
\usepackage{subcaption} % 用于创建子图和子图注
\usepackage{tikz} % 圆圈
\def\paperID{39583} % *** Enter the Paper ID here
\def\confName{CVPR}
\def\confYear{2026}

\title{SAR2Net: Learning Spatially Anchored Representations for Retrieval-Guided Cross-Stain Alignment}

\author{
Tianle Shen$^{1}$ \quad
Fang Yan$^{2}$ \quad
Xiaofan Zhang$^{1}$\thanks{Corresponding author: xiaofan.zhang@sjtu.edu.cn} \\
$^{1}$Shanghai Jiao Tong University \quad
$^{2}$Shanghai Artificial Intelligence Laboratory\\
}

\begin{document}
\maketitle
\begin{abstract}
% Achieving spatial alignment between whole-slide images (WSIs) across stains remains challenging due to extreme resolution, tissue fragmentation, and large nonlinear deformations. 
Achieving spatial alignment between whole-slide images (WSIs) across stains is essential for integrating morphology and molecular information, yet it remains challenging due to extreme resolution, tissue fragmentation, and large nonlinear deformations. 
Conventional registration pipelines depend on global pre-alignment and spatial consistency, which often collapse under such distortions.
We present SAR2Net, a framework that learns spatially anchored representations and reformulates cross-stain alignment as a region-level feature retrieval problem. 
Instead of estimating explicit transformations, SAR2Net learns pointwise representations encoding the relative spatial relationships to tissue landmarks. Given landmarks and arbitrary coordinates, it predicts spatially anchored features that serve as deformation-invariant descriptors of tissue topology. A two-stage retrieval framework then establishes correspondences between slides, even when global alignment is infeasible.
Experiments on biopsy-oriented HE-IHC datasets show that SAR2Net achieves robust region-level alignment under severe tissue distortions, outperforming previous methods. 
Code is available at \href{https://github.com/Shentl/SAR2Net}{https://github.com/Shentl/SAR2Net}.
% Code is available at {\small \url{https://github.com/Shentl/SAR2Net}}
% registration methods.
%outperforming both classical registration and recent deep-learning based methods.
% highly challenging

% Spatial alignment across histological stains in whole-slide images (WSIs) is notoriously difficult due to extreme resolution, tissue discontinuities, and large nonlinear deformations. Conventional registration pipelines depend on global pre-alignment and explicit transformation modeling, which often fail under such distortions.
% We introduce SAR²Net — Spatially Anchored Representation and Retrieval Network — which reformulates cross-stain alignment as a retrieval problem rather than a transformation estimation task. Instead of relying on human-defined landmarks, SAR²Net automatically discovers stable spatial anchors and learns pointwise representations that encode their relative topology. These spatially anchored features act as deformation-invariant descriptors, enabling correspondence retrieval across HE and IHC slides even when global alignment is infeasible.
% Extensive experiments on multiple HE–IHC datasets demonstrate that SAR²Net achieves robust region-level correspondence under severe tissue distortion, outperforming both classical and deep registration methods while providing interpretable spatial retrieval behavior.
% SAR²Net automatically identifies stable spatial anchors and learns pointwise representations encoding each location’s relative relationship to these anchors
\end{abstract}

% 那么你觉得我这篇CVPR投稿里的主图该怎么画，额外提供几点信息 1. 我的用于生成spatially anchored representations的模型是用对比学习训练的，不需要任何图像，全部用模拟的点做数据 2. landmarks用superpoint-superglue自动生成（这点不用体现在图上，只要体现有landmarks就行了）3. 我的retrieval框架现在缩略图下将HE划patch，然后选取patch周围的landmarks点对，计算HE中所有点和IHC中所有点在这些landmarks对下的spatially anchored representations，根据这些 representations进行retrieval, 如果IHC中某一个点的top_sim落在该HE patch中，那就认为该IHC点是HE patch对应的区域的一部分。这样retrieval一遍后，一个HE patch就对应一个初步的IHC 区域，然后就可以裁剪下并放大这对应区域，进行更细致retrieval    
\section{Introduction}
\label{sec:intro}
Histopathological examination is the gold standard for cancer diagnosis~\cite{cheng2023computational}. In pathology, tissue sections are stained using different dyes to reveal complementary biological information. Hematoxylin and eosin (HE) staining highlights tissue morphology, while immunohistochemistry (IHC) staining exposes molecular expression patterns. To jointly interpret morphological and molecular information, HE and IHC slides are typically prepared from adjacent sections of the same tissue, motivating the problem of cross-stain correspondence in computational pathology.

% HE and IHC slides of the same tissue must be spatially aligned, motivating the problem of cross-stain correspondence in computational pathology.
% Histopathological examination is the gold standard for cancer diagnosis. In pathology, tissue sections are stained using different dyes to reveal complementary biological information. Hematoxylin and eosin (HE) staining highlights tissue morphology, while immunohistochemistry (IHC) staining exposes molecular expression patterns. These stains are usually applied to adjacent tissue sections rather than the same one, making spatial correspondence between HE and IHC slides essential for joint interpretation and analysis.
% Histopathological examination is the gold standard for cancer diagnosis. While hematoxylin and eosin (HE) slides provide rich morphological cues, immunohistochemistry (IHC) slides reveal molecular expression patterns that are critical for accurate tumor typing and grading. To jointly interpret morphology and molecular signals, these slides must be spatially aligned, motivating the problem of cross-stain correspondence in computational pathology.
% Histopathological examination is the gold standard for cancer diagnosis. While hematoxylin and eosin (HE) slides provide rich morphological cues, immunohistochemistry (IHC) slides reveal molecular expression patterns that are critical for accurate tumor typing and grading. To jointly interpret morphology and molecular signals, these slides must be spatially aligned, motivating the problem of cross-stain correspondence in computational pathology.

Most existing approaches treat this problem as an image registration task, aiming to estimate a spatial transformation that aligns one slide with another. Given the gigapixel scale and stain heterogeneity of WSIs, registration is often performed in a multi-stage process, starting with coarse alignment at low magnification, followed by non-rigid refinement at higher resolutions. The methods typically rely on intensity-based optimization of similarity metrics (\eg MI, NCC, NGF), or feature descriptors matching. Recent works combine both strategies in hybrid pipelines~\cite{gatenbee2023virtual, wodzinski2024regwsi}, while others employ neural networks to directly predict transformation parameters~\cite{wodzinski2021deephistreg, roy2023deep}. Such methods with hierarchical designs perform well when a consistent global correspondence exists between slides.

\begin{figure}[t]
  \centering
   \includegraphics[width=1\linewidth]{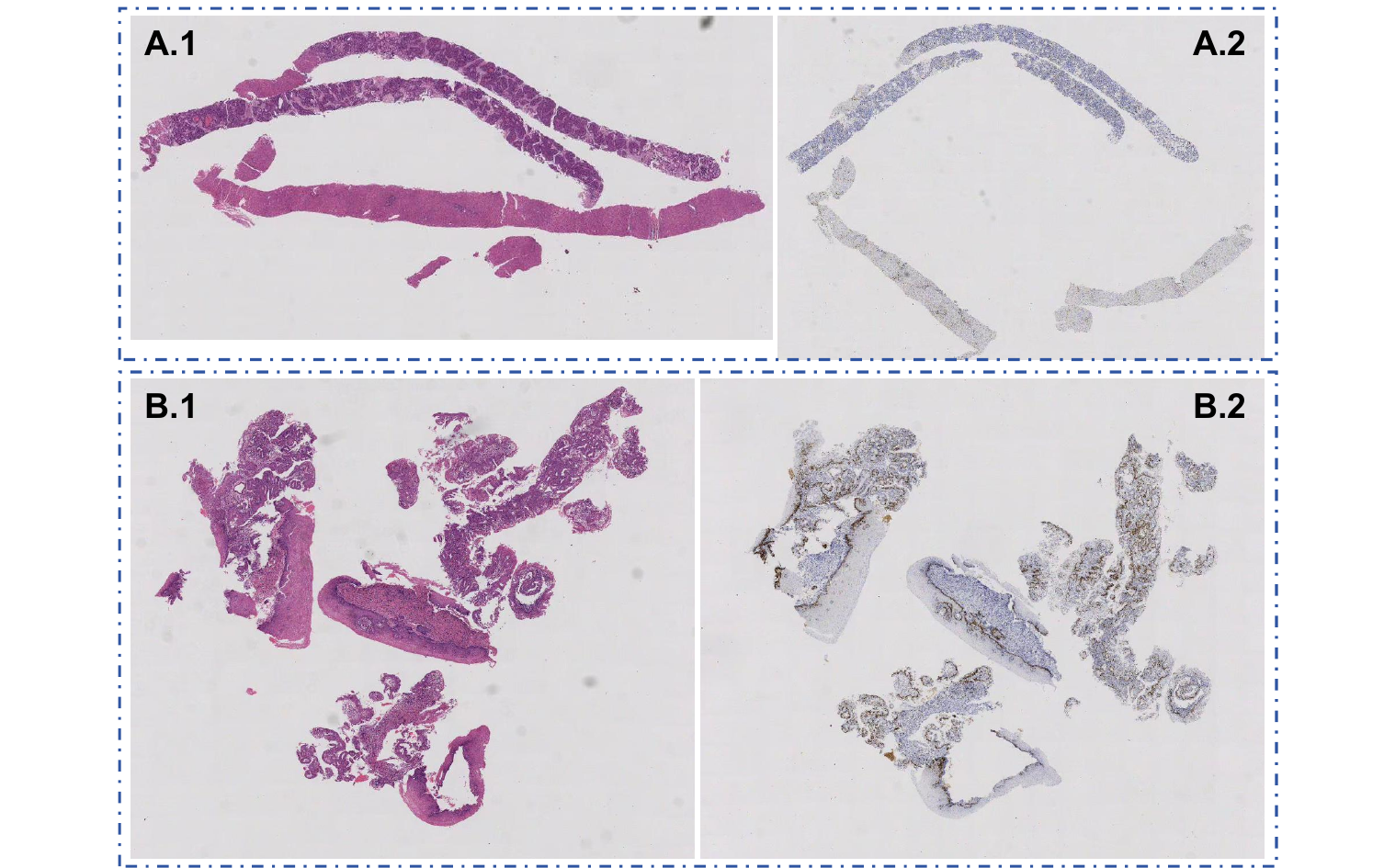}
   \caption{Examples of biopsy slides. (A.1-A.2) HE and IHC slides of a core needle biopsy with tissue fractures and bending. (B.1-B.2) slides of a forceps biopsy with inter-block displacement.}
   \label{fig:data_show}
\end{figure}

However, all these methods rely on the assumption that a coarse pre-alignment can be successfully established first.
This assumption often breaks down in histopathological slides with fragmented or heavily deformed tissue structures. In many biopsy-derived WSIs, tissues appear as multiple disconnected fragments or elongated strips that undergo bending, tearing, or nonlinear distortion during sectioning. As shown in \cref{fig:data_show}, adjacent HE–IHC slides of such samples frequently exhibit inconsistent spatial layouts, missing regions, or non-overlapping tissue parts, making global alignment unreliable. These complex structural discontinuities are common~\cite{abe2022development} in real-world pathology data but are rarely represented in existing benchmarks. When large portions of tissue cannot be coarsely aligned, conventional registration pipelines typically collapse, yielding unstable or meaningless correspondences.

% However, all these methods rely on the assumption that a coarse alignment can be successfully established. This assumption often fails for biopsy slides, where tissue discontinuities are inherent.  As shown in \cref{fig:data_show}, core needle biopsies are long and slender, prone to bending, distortion, or fragmentation during sectioning, while forceps biopsies consist of multiple small, separated tissue pieces that may drift, rotate, or disappear across adjacent sections. Such slides are ubiquitous~\cite{abe2022development} in clinical workflows due to their minimally invasive nature and diagnostic importance in early cancer detection and therapeutic planning~\cite{al2014whole,choi2024digital}, yet they remain largely overlooked in existing datasets and benchmarks. When large tissue fragments cannot be coarsely aligned, traditional registration pipelines tend to collapse, producing unstable or meaningless correspondences.

Interestingly, even when global alignment fails, local spatial relations often remain consistent within limited neighborhoods. Points that are close in the tissue tend to preserve their relative configuration despite large-scale deformation or tearing. This observation suggests that correspondence can be established by modeling point-wise relationships to nearby landmarks, rather than estimating a single global transformation. Furthermore, coarse region-level retrieval followed by local refinement allows the system to progressively narrow down corresponding areas, even under severe deformation.

Motivated by these insights, we introduce SAR2Net, a retrieval-guided framework that models geometric correspondence through learned spatial representations rather than explicit transformation prediction. SAR2Net encodes each point’s relative geometry with respect to automatically detected landmarks, which are treated as spatial anchors, forming representations that remain consistent across large cross-stain variations. Trained via contrastive learning on simulated coordinates without any image supervision, SAR2Net, coupled with a retrieval guided alignment pipeline, achieves robust region-level correspondence even under extreme deformations or tissue fragmentation, without requiring any pre-alignment.
% Motivated by these insights, we propose SAR2Net, a retrieval-guided framework for cross-stain alignment 
% by learning spatially anchored representations. 
% Instead of predicting transformation fields, SAR2Net learns spatially anchored representations—features that encode each point’s relative geometry to a set of automatically detected landmarks. These representations are trained via contrastive learning using only simulated point coordinates, without any image supervision. Combined with a multi-level retrieval pipeline, SAR2Net enables robust region-level correspondence across highly deformed or fragmented WSIs, without relying on coarse pre-alignment.

%Motivated by these observations, we introduce SAR2Net, a new paradigm for retrieval-guided cross-stain alignment. Instead of estimating transformation fields, SAR2Net learns spatially anchored representations: for any location in a slide, the model generates a feature encoding its relative spatial relationship to a set of learned landmarks. These features are robust to cross-stain differences and tissue distortions, enabling region-to-region retrieval between HE and IHC slides without requiring coarse pre-alignment. We further design a multi-level retrieval framework that progressively refines the correspondence from global to local scales.

Our contributions are threefold:
(1) We introduce SAR2Net, a network that models local geometric relations between spatial anchors and query points. The spatially anchored representations generated by SAR2Net achieve strong cross-stain robustness.
% propose a spatially anchored representation learning framework that models relative positional relationships rather than absolute coordinates, achieving strong cross-stain robustness.
(2) We develop a two-stage retrieval pipeline that can establish region-level correspondences under extreme tissue distortion or fragment misalignment.
(3) We demonstrate state-of-the-art performance on a biopsy-oriented cross-stain dataset, achieving reliable alignment in cases where traditional registration fails.

\section{Related Work}
\label{sec:related}

Traditional WSI registration methods can be broadly categorized into intensity-based and feature-based approaches~\cite{elhaminia2025traditional}.
Intensity-based methods directly exploit pixel-level information by optimizing similarity metrics like Mutual Information (MI) \cite{can2008multi,doyle2023whole}, Normalized Cross-Correlation (NCC) \cite{ma2008automatic,obando2017multi,kiemen2022coda}, and Normalized Gradient Fields (NGF) \cite{lotz2015patch,liang2021improving}. 
% These approaches capture global or local intensity discrepancies and perform well under rigid or mildly non-rigid transformations. % optimize global or local intensity consistency across images
In contrast, feature-based methods rely on the detection of salient local structures such as SIFT~\cite{paknezhad2020regional,shafique2021automatic,hoque2022whole} and SURF~\cite{rossetti2017dynamic,zhang2020point}, followed by descriptor computation and sparse correspondence estimation.
%between paired images, thus can handle complex deformations. 
However, due to strong stain variations, tissue deformations, and discontinuous tissue regions between WSI pairs, traditional methods often fail to achieve robust alignment.
To overcome the limitations of traditional methods, recent studies have employed deep learning models to extract more robust feature descriptors. CNN-based methods~\cite{awan2018deep,ge2022unsupervised,awan2023deep} learn discriminative representations from histopathological images and combine them with conventional matching algorithms for registration~\cite{boveiri2020medical}. SuperPoint~\cite{detone2018superpoint} further enables end-to-end keypoint detection and description, while SuperGlue~\cite{sarlin2020superglue} applies attention-based graph matching to establish reliable correspondences. The combination of SuperPoint and SuperGlue has been widely adopted in WSI registration. In addition, several pathology-specific approaches~\cite{liu2021histopathology,jeyasangar2024nuclei} detect nuclei as biologically meaningful keypoints to guide cross-stain alignment.

% 为克服传统方法在异质染色和复杂形变下特征匹配不稳定的问题，研究者开始利用深度学习模型来提取更具鲁棒性的特征描述符。一些研究\cite{xxx,xxx,xxx}采用卷积神经网络（CNN）从组织图像中学习特征表示，然后将这些学习到的特征用于后续的特征匹配和配准步骤，通常结合传统的匹配算法（如最近邻搜索或RANSAC）来进行几何估计。
% xxx 等人提出了SuperPoint \cite{detone2018superpoint} ，用于端到端地同时检测关键点并提取其特征描述符，显著优于SIFT等传统方法。随后，xxx等人提出了 SuperGlue \cite{sarlin2020superglue}，利用图神经网络基于注意力机制进行特征匹配，实现了更鲁棒的对应点建立。SuperPoint与SuperGlue的组合也因此被广泛应用于各种WSI配准算法中。
% 此外，在病理图像配准中，也有部分工作利用深度学习检测细胞核\cite{xxx,xxx}作为关键点，以此建立跨切片的生物结构对应关系。这类方法通过结合形态学信息与学习特征，使得在染色差异显著的 HE-IHC 配准中获得更具解释性和稳定性的匹配效果。

Beyond learning feature descriptors, some methods aim to directly estimate transformation parameters% using deep regression networks
, generating coarse affine alignments that are subsequently refined by non-rigid deformation models. For instance, Wodzinski \etal~\cite{wodzinski2021deephistreg} employed a ResNet~\cite{he2016deep} to predict affine transformation parameters, while Roy \etal~\cite{roy2023deep} generates multi-scale deformable vector fields using a FCN and additionally leverages a CycleGAN~\cite{zhu2017unpaired} to synthesize corresponding IHC slides from HE to assist registration. These parameter estimation approaches generally assume that slides can be roughly pre-aligned, which limits their applicability in cases with severe and complex tissue distortions.
% 除去利用深度学习模型提取特征描述符外，也有一些模型希望通过深度回归网络直接估计变换参数，生成粗略的仿射排列，然后通过非刚性变形模型进一步细化这些排列。Wodzinski 等人使用了 ResNet-101 网络进行仿射参数估计。Roy et al.提出的CGNReg通过一个Fully Convolutional Network生成多尺度的可变形矢量场来进行配准。该方法还使用CycleGAN从HE切片生成对应的IHC切片来辅助配准。这些参数估计的方法通常假设切片能通过初步配准的预处理基本对齐，限制了其在存在难以处理的强烈扭曲的切片上的应用

% 除去利用深度学习模型提取特征描述符外，其它一些工作还尝试利用生成模型从HE切片生成对应的IHC切片来辅助配准。也有一些模型希望直接估计变换参数，生成初步的仿射变换，或在基本配齐后进行非刚性的细节微调

% several studies have explored generative approaches to assist registration by synthesizing corresponding IHC-like images from HE slides, thereby reducing the cross-stain appearance gap~\cite{xxx,xxx}.

% Currently, the most robust WSI registration approaches often adopt hybrid strategies that combine robust feature matching, global optimization, and local deformation refinement. These methods typically employ multi-stage pipelines, applying different techniques at successive steps to handle both large-scale tissue alignment and fine-grained local variations. Representative examples include VALIS~\cite{gatenbee2023virtual} and RegWSI~\cite{wodzinski2024regwsi}, which integrate feature-based matching with global affine registration and subsequent non-rigid refinement, achieving stable performance across diverse HE-IHC slide pairs.

Currently, the most robust WSI registration approaches often adopt hybrid strategies that combine robust feature matching, global alignment, and local deformation refinement. 
% These multi-stage pipelines typically use SIFT, SuperPoint, or similar descriptors to perform initial registration, 
These multi-stage pipelines typically use descriptors like SuperPoint to perform initial registration, 
followed by non-rigid refinement. For example, after initial registration, VALIS~\cite{gatenbee2023virtual} conducts patch-wise optimization and obtains a dense deformation field by smoothly combining the local transformations, while RegWSI~\cite{wodzinski2024regwsi} performs non-rigid registration by optimizing local NCC at multiple scales. Such hybrid strategies leverage both structural correspondences and intensity cues, achieving stable and accurate alignment across heterogeneous HE-IHC slide pairs.

However, all existing methods rely on the assumption that WSIs can be roughly pre-aligned. 
% Whether using iterative optimization for non-rigid registration or regression-based parameter estimation, 
If large tissue regions in HE and IHC slides do not roughly overlap after initial affine alignment, subsequent refinement or learned models often lack a reliable starting point, leading to registration failure. % or clinically unacceptable errors. 
Such cases, caused by severe distortions or misaligned multi-tissue arrangements, are common in real-world pathological datasets, yet current research rarely address this challenge. In this work, we propose a method that circumvents the reliance on coarse pre-alignment, enabling robust alignment even under these difficult conditions.

\section{Methods}
\label{sec:methods}
\begin{figure*}[t]
  \centering
   \includegraphics[width=0.98\linewidth]{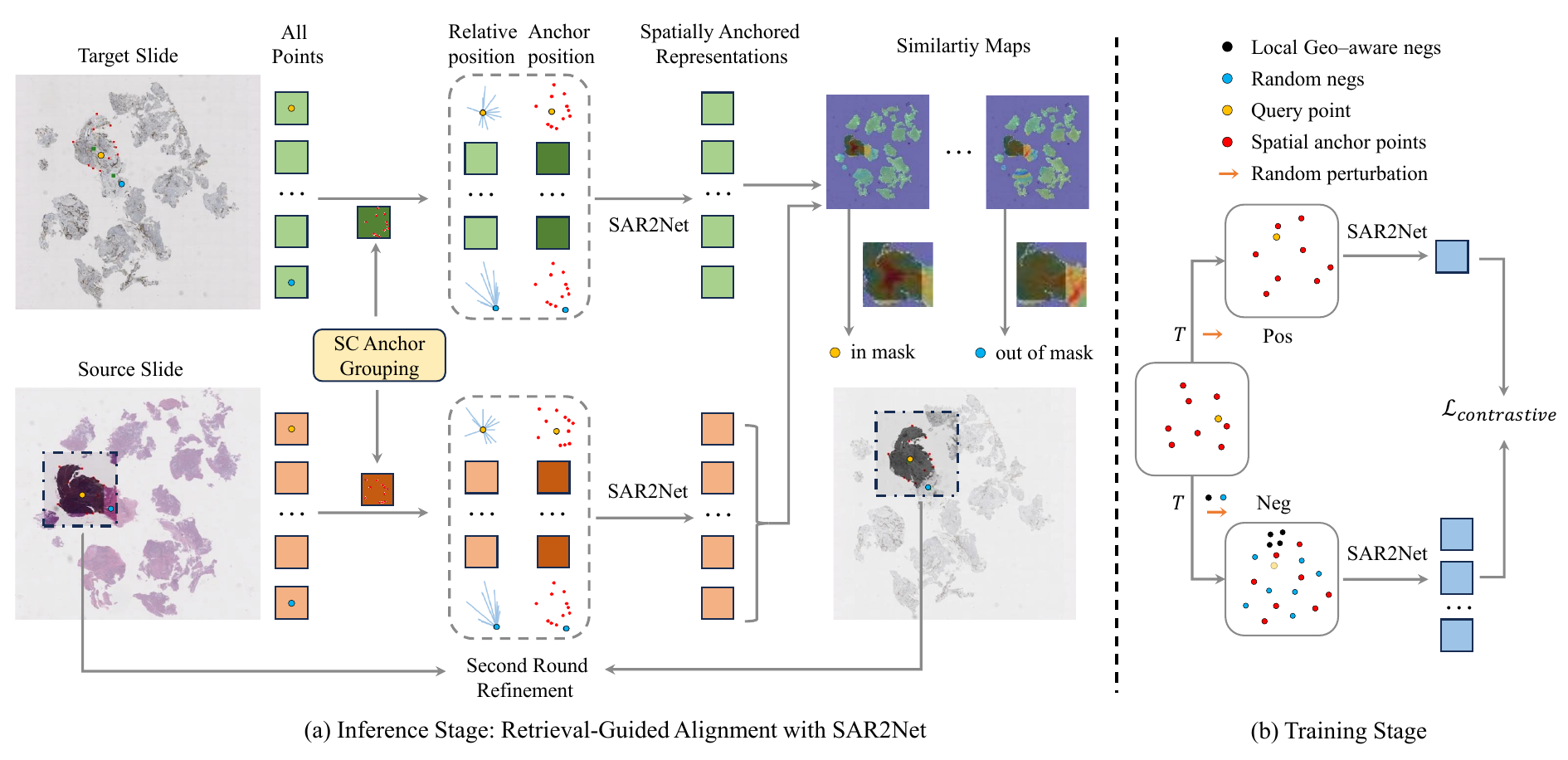}
   \caption{Pipeline of SAR2Net. (a) Given source and target WSIs, we extract paired landmarks automatically as spatial anchors. As detailed in Sec.~\ref{sec:retrieval}, for each source patch, nearby anchors are selected via Spatially Coherent Anchor Grouping. Then for all points in both slides, SAR2Net computes spatially anchored representations that encode their relative geometry to the anchors. 
   Afterwards, region correspondence is obtained through feature retrieval. If the highest similarity falls inside the source mask, the corresponding target points are assigned to the aligned mask. The retrieved regions are refined by a second round refinement.
   (b) As described in Sec.~\ref{sec:contrastive}, SAR2Net is trained via contrastive learning using purely synthetic point coordinates, combining random and local geometry–aware negative sampling.} % for all points in both slides
   \label{fig:pipeline}
\end{figure*}
% 都起个名字
% Our goal is to achieve robust cross-stain region alignment by modeling geometric relations directly from point correspondences, without relying on image appearance or pre-alignment.To this end, we propose SAR2Net, which learns spatially anchored representations $f(x, A)$ that encode the relative geometry between a query point $x$ and its surrounding anchors $A$.

Our goal is to achieve robust cross-stain region alignment, even under large global deformations or tissue distortions in WSIs.
To this end, we propose SAR2Net, a framework that learns spatially anchored representations capturing the local geometric relations between a query point $x \in \mathbb{R}^2$ and its surrounding anchors $A=\{a_i \in \mathbb{R}^2\}_{i=1}^K$, which remain consistent across slides. Our SAR2Net achieves robust region-level alignment across differently stained WSIs, without relying on image appearance or any pre-alignment.

In the following sections, we first describe the network architecture that constructs these representations (~\cref{sec:architecture}).
We then detail the contrastive training for SAR2Net with simulated points (~\cref{sec:contrastive}), followed by the retrieval guided region alignment pipeline that leverages these representations for cross-stain correspondence (~\cref{sec:retrieval}).

% 前面要提到anchor是pair的点对
\subsection{Model Architecture}
\label{sec:architecture}
% 介绍模型做什么、怎么输入输出、核心模块结构
% - 输入：坐标 + landmarks
% - 输出：spatially anchored feature
% - 网络：MLP / transformer-like
% 简洁、图示 + 简短公式
SAR2Net aims to generate deformation-robust features that encode each position’s local spatial relationship to a set of anchor points.
Given a variable number of anchors $A=\{a_{i} \in \mathbb{R}^2\}_{1=1}^{K}$, whose spatial configuration is assumed to remain roughly consistent across paired slides, and a query point $x \in \mathbb{R}^2$, we first compute their relative positions $R=\{r_{i} | r_{i}=x-a_{i}, i=1,...,K\}$, and concatenate each $r_{i}$ with its corresponding anchor coordinate $a_{i}$ to form a position descriptor:
\begin{align}
    p_{i} = [r_{i}, a_{i}] \in \mathbb{R}^{4}.
\end{align}
% 写出 a 代表anchor的空间分布，r表示到anchor的空间分布
All descriptors $p_{i}$ are passed through a MLP $\phi_{\theta}$ that encodes local spatial relations into embeddings $h_{i}$ with $d=256$: % All descriptors $p_{i}$ ... into latent embeddings $h_{i}$ ...
\begin{align}
    h_{i} = \phi_{\theta}(p_{i}) \in \mathbb{R}^{d}.
\end{align}

To encode the global spatial position of an arbitrary query point relative to a set of anchors, we aggregate spatial information from multiple anchors. Each anchor contributes differently based on its geometric relation to the query point: anchors closer to the query capture fine-grained local variations, while distant anchors provide stable global constraints. Therefore, the resulting spatially anchored representation of point $x$ can be formulated as a weighted spatial embedding within the anchor set: 
%To characterize the global spatial position of a query point within the anchor set, we aggregate spatial information from multiple anchors.Considering that anchors at different distances contribute differently—nearby anchors are more sensitive to local structural variations, while distant anchors provide more stable global spatial constraints—the final spatially anchored representation can be interpreted as a weighted projection of the query in the anchor space.
\begin{align}
    f(x,A) = \sum_{i=1}^{K} \alpha_{i}h_{i},
\end{align}
\begin{align}
    \alpha_{i} &= \frac{\exp \left\{ W\left(\tanh(V h_{i}) \odot \text{sigm} (U h_{i})\right)\right\} }{\sum_{j=1}^{K} \exp \left\{ W \left(\tanh(V h_{j}) \odot \text{sigm} (U h_{j})\right) \right\}},
\end{align}
where $f(\cdot, \cdot)$ is our SAR2Net parameterized by $U,V,W$, $\odot$ denotes element-wise multiplication.
% The final spatially anchored representation of point $x$ is obtained as a weighted aggregation of its anchor-level embeddings:

\subsection{Contrastive Learning of Spatial Geometry}
\label{sec:contrastive}
SAR2Net is trained to learn spatially anchored representations that depend solely on point coordinates rather than image appearance, encoding the geometric relations between a query point and its surrounding anchors.
Given $K$ anchors $A=\{a_i\}_{i=1}^K$ and a query point $x$, the network generates a representation $f(x, A) \in \mathbb{R}^d$ that captures the query’s relative spatial configuration within the anchor set.
% SAR2Net 旨在仅根据查询点与其周围锚点的位置关系，学习到一个能够稳定表征局部几何结构的空间锚定表示，而无需任何图像特征。换言之，给定一组锚点坐标 ${a_i \in \mathbb{R}^2}_{i=1}^{n}$ 和任意查询点 $q \in \mathbb{R}^2$，网络通过建模查询点与锚点集合的相对空间几何关系，输出一个特征向量 $f(q, {a_i}) \in \mathbb{R}^d$，该特征在跨染色、局部形变下保持一致。

% using only simulated point coordinates
To learn such spatial representations, we adopt a contrastive training scheme using only simulated point coordinates. In each training iteration, we randomly sample $K+1$ points, where $K$ points form the anchor set $A$ and the left one point serves as the query $x$. A rigid transformation $T(\cdot)$ is applied to generate a positive pair $\left((x, A), (T(x), T(A))\right)$, reflecting the same spatial structure.
% under different coordinate systems. 
Negatives are constructed by applying $T(\cdot)$ to the same anchor set $A$ and resampling $N$ negative query points $\{x_j^{-}\}_{j=1}^{N}$ that satisfy $\|x_j^{-} - T(x)\| > \delta_{neg}$, ensuring sufficient geometric separation.

\begin{figure}[t]
  \centering
   \includegraphics[width=1\linewidth]{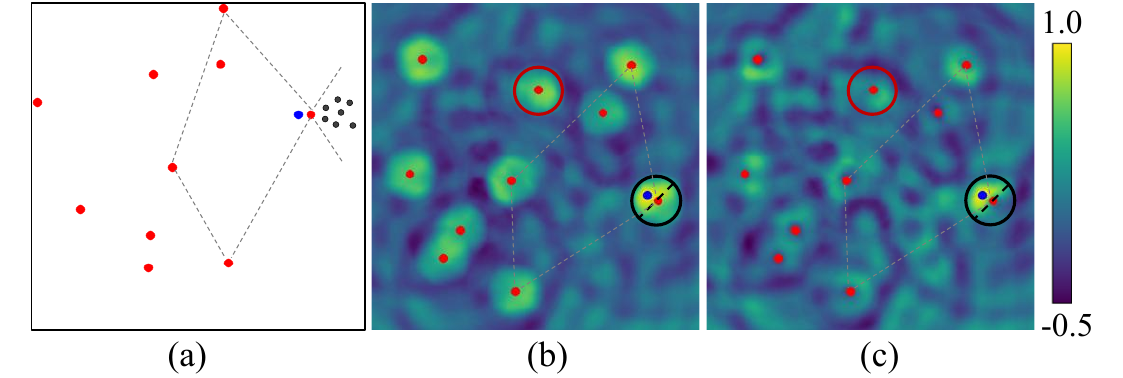}
   \caption{Local geometry–aware negative sampling. (a) Example of a query (blue) near anchors (red), with LGN samples (black). (b) Sim. map w/o LGN. (c) Sim. map w/ our convex-hull based LGN, enabling the model to capture local geometric relations. In (b-c), the \textbf{black circle} highlights a local neighborhood around the reference point that is very close to the query, while the \textbf{red circle} marks a neighborhood around an anchor that is far from the query.} 
   \label{fig:train}
\end{figure}

When the query lies very close to an anchor, the model may focus excessively on anchor’s local neighborhood and neglect the overall anchor geometry. To address this, we employ \textbf{local geometry–aware negative sampling (LGN)}. As illustrated in \cref{fig:train} (a), the nearest anchor is treated as a reference, and a local convex hull is formed with its neighboring anchors. Negative queries are then sampled symmetrically on the opposite side of the hull, maintaining similar distances to the reference. In \cref{fig:train} (b), without LGN, the model tends to produce high similarity around both the reference anchor (black circle) and other anchors (red circle), effectively learning only that the query is near an anchor. 
% In \cref{fig:train} (c), with LGN, the high-similarity region in the black circle is constrained within the convex hull while similarities around other anchors are suppressed, encouraging the model to capture richer local spatial geometry.
In \cref{fig:train} (c), with LGN, the high-similarity region in the black circle is constrained within the convex hull, while the similarity around the remaining anchors in the red circle is also suppressed.
% while similarities around other anchors in the red circle
%, encouraging the model to capture richer local spatial geometry.
At this point, the model captures more local spatial geometric information. % learns

% such as whether a query lies inside or outside the local geometric structure. %local anchor structure. % is constrained to the interior of the convex hull

% As illustrated in \cref{fig:train_r}, without LGN the model tends to produce high similarity around both the reference anchor and other nearby anchors, effectively learning only that the query is near an anchor. With LGN, the high-similarity region is constrained to the interior of the convex hull while similarities around other anchors are suppressed, encouraging the model to capture richer local spatial geometry, such as whether a query lies inside or outside the local anchor structure.

% rebuttal text: In \cref{fig:train_r} (b), when the query is very close to an anchor (as the reference point), both the similarity around the reference point (black circle) and that around the remaining anchors (red circle) are high. The model overfits to absolute proximity, merely learning ``query is near an anchor.'' However, with local geometry-aware negative sampling, in \cref{fig:train_r} (c), the high-similarity region in the black circle is restricted to the inside of the convex hull formed by the reference point and surrounding anchor points, while the similarity around the remaining anchors in the red circle is also suppressed. At this point, the model learns more local spatial geometric information.

Finally, a random uniform perturbation $\epsilon \sim \mathcal{U}(-\sigma, \sigma)$ is added to all coordinates to improve robustness and spatial smoothness, which simulates the subtle nonlinear deformations commonly observed in real tissue sections. The network is trained with a standard InfoNCE loss.

% train4.pdf \caption{Local geometry–aware negative sampling. (a) Example of a query (blue) located near anchors (red), with local geometry–aware negative samples (black). (b) Similarity map without our geometry-aware sampling. (c) Similarity map with our convex-hull–based negative sampling, enabling the model to capture local geometric relations.} 

% 强调创新训练机制（纯几何、无图像）
% - 数据模拟方式
% - 正负样本定义
% - InfoNCE 目标
% 带点公式，突出思想“learn from geometry only”
\subsection{Retrieval-Guided Region Alignment}
\label{sec:retrieval}
% 说明如何用这些特征实现跨染色对齐
% - 多级检索策略
% - 相似度计算
% - 区域映射逻辑
% 流程性强，可配小流程图
\paragraph{Spatial Anchor Extraction}
% cleaning的逻辑先不写 (不是主要)
To obtain spatially anchored representations, reliable spatial anchors are required as geometric reference points across stains.
% In this work, 
We treat paired landmarks from source and target images as spatial anchors. We extract them by directly applying pre-trained SuperPoint and SuperGlue models on low-resolution slide thumbnails.
% To extract these anchor points, we direct apply pre-trained SuperPoint and SuperGlue models on slide thumbnails. 
The SuperPoint-SuperGlue framework is known for its excellent generalization ability, enabling robust correspondence across different histopathological slides.

However, both SuperPoint and SuperGlue lack rotation invariance, while tissue sections in WSIs may undergo various rotations. To address this, we rotate the source image by multiple angles and extract anchor points for each rotation. The rotation that yields the largest number of valid correspondences is selected, and its matched keypoints are used as the main anchor set $\left(A_{source}, A_{target}\right)$, where $A_{source}=\{a_{i}^{s}\}_{i=1}^{K}$ and $A_{target}=\{a_{i}^{t}\}_{i=1}^{K}$.

Subsequently, anchor point pairs $(a_{j}^{s}, a_{j}^{s})$ extracted from other rotation angles are added to the main anchor set if for all $i \in N$, $\max(\| a_{j}^{s} - a_{i}^{s} \|, \| a_{j}^{t} - a_{i}^{t} \|)< \delta_{merge}$. 
%and $ \| a_{j}^{t} - a_{i}^{t} \|< \delta$, where $\delta$ is a small distance threshold for anchor merging. 
This step ensures that when multiple regions or significant distortions are present in the WSI, spatial anchors in different local areas with varying rotations are extracted accurately. % (such as fragmentation or bending)
% accurately -> 完备的

\paragraph{Spatially Coherent Anchor Grouping}
% Localized Geometric Consistency Filtering
% Local Consistency-Aware Anchor Grouping
% Local Geometry-Constrained Anchor Selection

Our region-level alignment assumes that the spatial relationship between a query and its nearby anchors remains locally stable across stains. However, large tissue deformations and discontinuities in WSIs often break this assumption, making global anchor sets unreliable for correspondence.

To preserve spatial validity, we adopt a spatially coherent anchor grouping strategy. For each region to be aligned, we slide a window over the source slide and select up to $K$ nearest anchors within the window and its surrounding neighborhood, ensuring locally valid geometric relations.

Still, anchors within the same window may not form a coherent structure, as some are mismatched, while others arise from separate tissue fragments or deformation patterns. To refine the grouping, we apply iterative geometric consensus using RANSAC~\cite{fischler1981random} to cluster anchors that share a compatible local transformation, progressively discarding outliers until no valid model remains under the threshold $\delta_{split}$. Each resulting subset serves as a stable geometric basis for subsequent region-level alignment.

\paragraph{Region Retrieval and Alignment}
For each window, we compute spatially anchored representations for all points in both slides using same anchors. Region correspondence is then established through feature retrieval: for each target point, we map its similarity against all points in source slide and identify its top-$k$ most similar source points. If these points are concentrated within the corresponding source window mask, the target point is assigned to that region. % same set of nearby anchors % For each local window % using the same set of anchors % we calculate its similarity map

However, in some cases, the target region may have no true counterpart in the source slide due to tissue loss, fragmentation, or severe deformation. In such cases, the top-$k$ matches may still fall within nearby source areas but exhibit uniformly low similarity, leading to spurious correspondences. To prevent this, we introduce a similarity threshold $\delta_{sim}$ and only assign a region label when the maximum similarity exceeds this value. This filtering step effectively suppresses false matches and enhances the reliability of cross-stain region alignment.

% \Large \huge \Huge
\begin{figure}[t]
  \centering
   \includegraphics[width=1\linewidth]{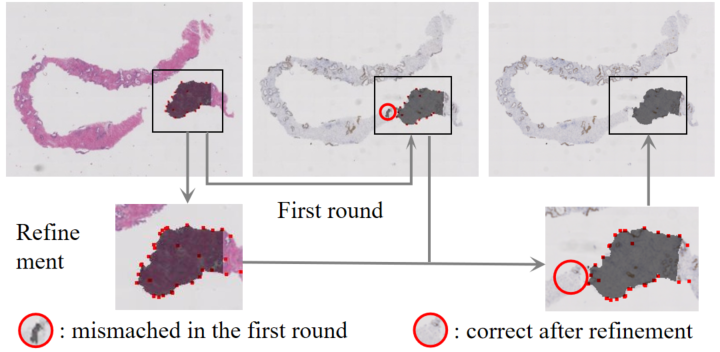}
   \caption{Two-Stage Refinement. The second-stage refinement is operated on each region and its neighborhood obtained from the first-stage retrieval. The \textcolor{red}{\huge$\circ$} indicates mismatched areas from the first stage that are successfully corrected after refinement.
  }
   \label{fig:refinement}
\end{figure}
% 第二轮Refinement在第一阶段得到的区域及其邻域上进行，其中红色圆圈代表了在第一轮retrieval中错配的区域，在更二轮优化中得到了更正

\paragraph{Two-Stage Refinement}
After the first round of retrieval, our model has already established a reliable region-level correspondence. However, due to the large window size and sparse anchor distribution, local areas may still exhibit slight nonlinear deformations or boundary shifts, limiting alignment precision. To further refine local correspondence, we introduce a two-stage refinement strategy. Specifically, as illustrated in \cref{fig:refinement}, each matched region from the first stage is treated as a new region of interest (ROI), within which additional anchors are extracted to form denser and more locally consistent anchor sets. The region alignment process is then re-applied within these refined ROIs. Built upon the initial region-level mapping, this hierarchical refinement progressively enhances alignment accuracy and stability under complex tissue deformations.
\section{Dataset}
\label{sec:dataset}

\subsection{Data collection and cleaning}
% 时态改为现在式节省空间
% We collected a subset of histopathology slides
% We collected histopathology slides from Xijing Hospital. Approximately 35$\%$ of these slides exhibited severe tissue distortions or consisted of multiple disjoint tissue fragments, characteristics commonly observed in biopsy specimens. From the raw collection, we selected 154 cases that contained both HE and IHC slides, ensuring that at least part of the tissue regions in each HE slide had corresponding areas in the IHC counterpart. In some cases, one HE slide is associated with multiple IHC slides (up to over ten). For cases with more than three corresponding IHC slides, we randomly selected up to three to maintain data diversity and computational feasibility. In total, we obtained 370 HE–IHC slide pairs covering 79 different IHC staining types. Our dataset spans diverse tissues and cancer types, with 23.6\% from breast, 15.7\% from liver, and 14.6\% from prostate. % Our dataset is pan-tissue and pan-cancer
We collect histopathology slides from Xijing Hospital. Approximately 35$\%$ of these slides exhibit severe tissue distortions or consisted of multiple disjoint tissue fragments, characteristics commonly observed in biopsy specimens. From the raw collection, we select 154 cases that contain both HE and IHC slides, ensuring that at least part of the tissue regions in each HE slide have corresponding areas in the IHC counterpart. In some cases, one HE slide is associated with multiple IHC slides (up to over ten). For cases with more than three corresponding IHC slides, we randomly select up to three to maintain data diversity and computational feasibility. In total, we obtain 370 HE–IHC slide pairs covering 79 different IHC staining types. Our dataset spans diverse tissues and cancer types, with 23.6\% from breast, 15.7\% from liver, and 14.6\% from prostate. % Our dataset is pan-tissue and pan-cancer

% num patients 154
% num pairs 381
% num stains 81
\begin{figure}[t]
  \centering
   \includegraphics[width=1\linewidth]{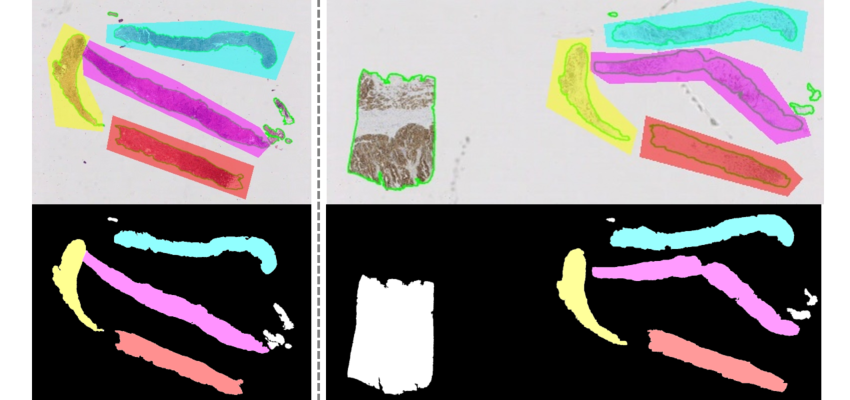}
   \caption{Example of data annotation. The left and right slides are adjacent slides, where the right one exhibits severe distortion and has an additional control tissue. After segmenting the tissue foreground (green solid lines) and annotating the corresponding regions, four paired region masks were obtained.} % Among them, the second mask exhibits severe distortion due to tissue tearing and folding.
   \label{fig:data_anno}
\end{figure}
\subsection{Data Annotation} % Region-level annotation
% 注释

Considering that region-level alignment is often sufficient for diagnostic interpretation, we annotated our 370 HE–IHC pairs at the region level. We first performed automatic tissue–background segmentation to accelerate the subsequent annotation process. The segmentation was conducted using the tissue segmentation module from Trident~\cite{zhang2025accelerating}, which is robust across various staining types and does not require manual parameter tuning as in traditional threshold-based approaches. After segmentation, we manually delineated corresponding tissue regions, ensuring that each annotated region was complete and consistently present in both HE and IHC slides. \cref{fig:data_anno} illustrates examples of the annotated region correspondences.
% 无需精确到细胞级，甚至说对于连续多张切片，某张切片上有的结构到后面几张就没有了

% 考虑到病理医生在实际判读时很多时候只需要做到区域级别的对齐，我们对我们收集的来的381对HE-IHC 全切片对进行了Region级别的标注。首先，我们进行了全自动的前景分割，这能大大加速后续的区域标注。我们使用 Trident中的组织分割模块进行组织与背景的分割，该模块对不同的染色剂具有鲁棒性，并且不需要像以前的阈值方法那样进行手动调整。然后，我们在分割出的前景上标出了不同的组织块，我们确保标注的组织块在HE和对应的IHC上都存在且完整，如图xxx所示。
% we use the tissue segmentation module in Trident~\cite{zhang2025accelerating} to do tissue-vs-background segmentation, which is robust to different stains and do not require manual tuning like the previous thresholding methods.
% 放示意图

\section{Experiments and Results}
\label{sec:experiments}
% \paragraph
\subsection{Implementation Details}
% neg_samples N
% batch_size
% num_anchors
% $\delta_{neg}$
% perturbation_parameter $\sigma$
During training, we use simulated 2D coordinates to enable SAR2Net to learn spatial relationships between query and anchor points. The output embedding dimension $d$ is 256. With a batch size of 256, the number of spatial anchors $K$ is fixed within a batch but randomly sampled from $\{4, \dots, 10\}$ across batches. For negative sampling, we set the minimum separation distance $\delta_{neg}=20$ and draw $N=500$ negative queries for each sample. To encourage the network to reason about local geometric boundaries, 10$\%$ of query points are sampled near anchors, for which additional negatives are placed symmetrically on the opposite side of the convex hull, as described in Sec.~\ref{sec:contrastive}. For all sampled coordinates, a uniform perturbation $\epsilon \sim \mathcal{U}(-5, 5)$ is added to improve robustness. The model is trained with the Adam optimizer using a learning rate of $1\times10^{-4}$ and the InfoNCE loss~\cite{oord2018representation} with temperature $\tau=0.3$.

During inference, spatial anchors are extracted using SuperPoint and SuperGlue with default settings.
The source slide is rotated every $15^\circ$, and anchors detected across all rotations are merged with $\delta_{merge} = 5$ and split with $\delta_{split} = 10$.
Region retrieval is then performed with sliding windows of size $100$ and stride $50$, each associated with up to $15$ nearest anchors (or all available if $K_{anchors} < 15$ in that area).
A target point is assigned to a source region if any of its top-$5$ most similar source points lie within the region mask and the similarity exceeds $\delta_{sim}=0.7$.

% During inference, we extract spatial anchors using SuperPoint and SuperGlue with their default configurations, and merge nearby matches within a distance threshold of $\delta_{\text{merge}} = 5$. For region-level retrieval, we slide windows of size $100$ with a stride of $50$ over the source thumbnail. Each window uses up to $K = 15$ nearest anchors in its neighborhood; if fewer anchors are available, all $k_{\text{anchors}} < 15$ anchors are used instead. When performing spatial mapping with the spatially anchored representations produced by SAR2Net, a target point is considered to belong to a source mask if its top-$5$ most similar source points fall within that mask.
% 在我们的训练阶段，我们使用模拟的坐标点来使我们的SAR2Net能够学习查询点和锚点点之间的空间关系，模型输出的Spatially Anchored Representations维度为256。我们使用batch_size为256，一个batch中，采样的空间锚点的数量K保持不变，而在不同batch间，K在[4,10]的整数中随机采样。为了采样负样本对，我们设置$\delta_{neg}$ 为20，在每条数据中，负样本对的数量N为500。我们以0.1的比率采样那些靠近空间锚点的查询点，并在构建凸包后专门在对侧采样负样本点，同时保持全局的负样本点采样，就像在section{Contrastive Learning of Spatial Geometry }中提到的那样。针对每一个采样点，perturbation parameter $\sigma$ 为5。训练时，我们使用Adam优化器，学习率为1e-4。我们使用InfoNCE Loss，温度$\tao$为0.3

% 在推理阶段，我们使用SuperPoint和SuperGlue默认的参数提取空间锚点，并设置$\delta_{merge}$为5。对每个sliding window，我们设置patch_size为100，stride为50，并选取该sliding window周围K=15个最近的锚点，如果周围锚点的数量$k_{anchors}$小于15，那么$K=k_{anchors}$。在使用SAR2Net提取的spatially anchored representations进行空间映射时，top-5的被使用来判断target point是否在mask中
 
\begin{table}[t]
    \caption{Quantitative performance with mIoU, mDice, aw-IoU, and aw-Dice. Ours (single) means only the first round region retrieval and alignment is performed. Ours (second) means the second round refinement is performed.} % Quantitative performance with mIoU, mDice, aw-IoU, and aw-Dice. Ours (single) means only the first round region retrieval and alignment is performed. Ours (second) means the second round refinement is performed. % Quantitative performance with mIoU, mDice, aw-IoU, and aw-Dice. Ours (single): only the first round region retrieval and alignment; Ours (second): the second round refinement is performed.
    \centering
    \small
    \begin{tabular}{lcccccccc}
    \toprule
     Method & mIoU $\uparrow$ & mDice $\uparrow$& aw-IoU $\uparrow$& aw-Dice $\uparrow$\\
   \midrule
    VALIS~\cite{gatenbee2023virtual}  & 0.635 & 0.705 & 0.647 & 0.716 \\
    RegWSI~\cite{wodzinski2024regwsi} & 0.691 & 0.786 & 0.699 & 0.794 \\
    Ours (single)  & \underline{0.891} & \underline{0.938} & \underline{0.893} & \underline{0.939} \\
    Ours (second)   & \textbf{0.899} & \textbf{0.942} & \textbf{0.901} & \textbf{0.944} \\
    \bottomrule
    \end{tabular}
    \label{tab:overlap}
\end{table}

\subsection{Metrics}
To systematically evaluate the region-alignment performance, we report the mean intersection over union (mIoU) and mean dice similarity coefficient (mDice). To mitigate bias from different region sizes, we also report area-weighted versions (aw-IoU, aw-Dice), where the metrics are averaged by region area before taking the mean. 

In addition, we assess robustness using the success rate $\mathrm{SR}_{\mathrm{IoU}}^{t}$ and $\mathrm{SR}_{\mathrm{Dice}}^{t}$, defined as the proportion of regions whose IoU or Dice exceeds a threshold $t$:
\begin{align}
    \mathrm{SR}_{\mathrm{Metric}}^{t} = \frac{1}{M}\sum_{k=1}^{M}\mathbf{1}\left[\mathrm{Metric}_k \ge t\right],
\end{align}
where $\text{Metric}_k$ denotes IoU or Dice of region~$k$. The area under the success-rate curve (AUC) is also reported.
% While mean-based metrics reflect average performance, the success rate characterizes the proportion of well-aligned regions above a given threshold, offering a complementary view of robustness.
% While the mean-based metrics reflect overall accuracy, the success rate highlights the proportion of regions achieving high-quality alignment, thereby providing a more distribution-aware measure of robustness.

While mean-based metrics reflect average alignment accuracy, the success rate characterizes the proportion of well-aligned regions above varying thresholds, providing a more distribution-aware and robustness-oriented evaluation.

\subsection{Baseline} We compare our SAR2Net with two recent leading WSI alignment frameworks: VALIS~\cite{gatenbee2023virtual} and RegWSI~\cite{wodzinski2024regwsi}. Both VALIS and RegWSI are fully automated, multi-stage WSI registration frameworks that perform feature-based rigid or affine initialization followed by intensity-based non-rigid refinement. We evaluate all methods on our biopsy dataset to assess their performance under severe tissue discontinuities and deformations.

\begin{table}[t]
    \caption{Robustness Analysis with success rate (SR) over IoU and Dice. Ours (single) means only perform the first round region retrieval and alignment. Ours (second) represents that the second round refinement is performed.}
    \centering
    \small
    \begin{tabular}{lcccc}
    \toprule
      &  $\mathrm{SR}_{\mathrm{IoU}}^{0.75} \uparrow$ & $\mathrm{SR}_{\mathrm{IoU}}^{0.85} \uparrow$ & $\mathrm{SR}_{\mathrm{Dice}}^{0.75} \uparrow$ & $\mathrm{SR}_{\mathrm{Dice}}^{0.85} \uparrow$ \\
   \midrule
    VALIS~\cite{gatenbee2023virtual}  & 0.594 & 0.429 & 0.690 & 0.606\\
    RegWSI~\cite{wodzinski2024regwsi} & 0.545 & 0.324 & 0.740 & 0.574 \\
    Ours (single)  & \underline{0.899} & \underline{0.768} & \underline{0.971} & \underline{0.910} \\
    Ours (second)  &\textbf{ 0.939} & \textbf{0.812} & \textbf{0.974} & \textbf{0.946}\\
    \bottomrule
    \end{tabular}
    \label{tab:sr_main}
\end{table}

\begin{figure}[t]
  \centering
   \includegraphics[width=1\linewidth]{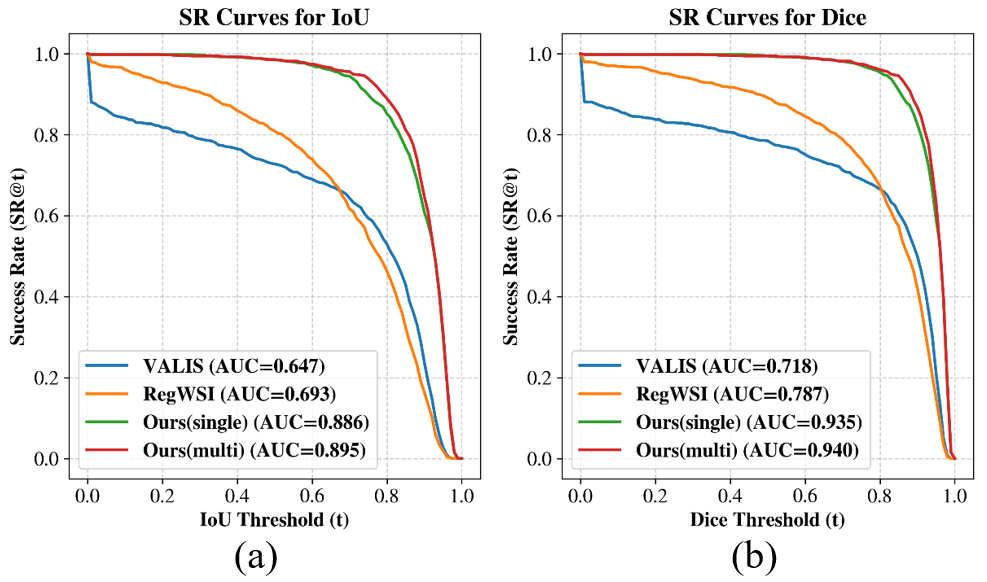}
   \caption{Visualization of the success rate curves. (a) Success rate curves over IoU thresholds. (b) Success rate curves over Dice thresholds. The AUC of the curves are reported on the figures.} % Evaluation of the region alignment performance with success rate.
   \label{fig:sr_main}
\end{figure}

\subsection{Quantitative Comparisons}
% , especially under the large nonrigid deformations commonly observed in whole-slide images.
As shown in \cref{tab:overlap}, our method achieves substantial improvements over existing approaches across all four metrics (mIoU, mDice, aw-IoU, and aw-Dice). This demonstrates that the proposed alignment framework is considerably more effective than traditional registration pipelines. Moreover, incorporating the second-round refinement yields additional and consistent gains over the single-round variant, confirming the effectiveness of our iterative retrieval-and-refinement strategy for correcting residual misalignments.
% 如表~\ref{tab:overlap}所示，我们的方法在所有四个指标（mIoU、mDice、aw-IoU 和 aw-Dice）上均显著优于现有方法。这表明，我们提出的配准框架比传统的配准流程更加有效。此外，引入第二轮优化后，性能较单轮优化方案得到了进一步的稳定提升，证实了我们迭代检索和优化策略在校正残余错位方面的有效性。

% 与基于均值的指标相比，面积加权指标（aw-IoU 和 aw-Dice）更强调较大区域的贡献，因此对区域大小分布不均更为敏感。对于 Valis 和 RegWSI 等先前的方法，尽管它们的 aw 指标略高于其均值指标，但这种改进主要由少数较大且易于配准的区域所驱动，而许多中等和较小的区域仍然配准不佳。这反映了它们对精确初始比对的强烈依赖性，以及当几何结构局部变得复杂时鲁棒性不足的问题。
%Compared with mean-based metrics, the area-weighted counterparts (aw-IoU and aw-Dice) emphasize the contribution of larger regions and are therefore more sensitive to uneven region-size distributions. For previous methods such as Valis and RegWSI, although their aw-metrics are slightly higher than their mean metrics, this improvement is largely driven by a few large and easier-to-align regions, while many medium and small regions remain poorly aligned. This reflects their strong dependence on an accurate initial alignment and their limited robustness when the geometric structure becomes locally complex.
For previous methods such as VALIS and RegWSI, the area weighted metrics are slightly higher than the mean metrics, which is mainly driven by the correct alignment of a few large and easier-to-match regions. This indicates an imbalance in their alignment behavior: while large regions are aligned reasonably well, many medium and small regions remain poorly aligned.
Such imbalance stems from the tendency of these registration frameworks to rely heavily on the initial alignment stage, which often prioritizes the global fit of larger regions at the expense of smaller ones. As a result, these methods struggle to maintain robustness once the geometric structure becomes locally complex.

In contrast, our method still exhibits a markedly larger advantage on the area weighted metrics and, more importantly, shows only minimal discrepancy between mean based metrics and area weighted metrics. This indicates that our alignment quality is uniformly high across regions of different sizes, rather than being biased toward large regions. Such closeness between the mean and area weighted metrics highlights the size-robustness of our framework and demonstrates that it maintains stable and reliable alignment performance for both large and small regions.

\subsection{Robustness Analysis with Success Rate}

As shown in \cref{tab:sr_main}, our method achieves substantially higher success rates than existing approaches across different IoU and Dice thresholds. This indicates that, compared with previous registration pipelines, our retrieval–alignment framework maintains a much larger proportion of high-quality aligned regions, yielding more stable and robust performance overall. Moreover, the second-round refinement brings additional improvements across all SR metrics, particularly at higher thresholds, demonstrating its effectiveness in correcting the residual misalignments left by the first round and improving performance in difficult regions.

This trend is further corroborated by the success rate curves in \cref{fig:sr_main}. The AUC values shown in the figure reveal that our method consistently maintains higher SR curves across the entire range of thresholds, highlighting its advantage across regions of varying difficulty. In contrast, VALIS and RegWSI exhibit highly unstable behavior when reliable initial alignment is unavailable: many regions fall below even moderate IoU/Dice thresholds, and VALIS even collapses entirely on certain samples, underscoring the vulnerability of previous frameworks under large deformations and complex local structures. The gains from our second round refinement concentrate in the high threshold ranges, 0.7-0.9 for IoU and 0.8-0.9 for Dice, which aligns well with our design: the first round retrieval in SAR2Net establishes a strong global correspondence, while the second round refinement corrects fine-scale nonlinear distortions and boundary shifts remaining from the first round, thus increasing the proportion of high-quality matches.

Together, \cref{tab:sr_main} and \cref{fig:sr_main} demonstrate that our method not only achieves superior average overlap performance, but also delivers a more stable and consistent alignment quality distribution. It maintains reliable performance across regions of different sizes and difficulty levels, significantly outperforming existing WSI registration approaches.

\begin{table}[t]
    \caption{Ablation study on the presence of random perturbations (pertu), local geometry-aware negative sampling (LGN), and second-round refinement (sec).}
    \centering
    \small
    \begin{tabular}{cccccc}
    \toprule
    w. pertu & w. LGN  & w. sec & mIoU $\uparrow$ & mDice $\uparrow$& aw-IoU $\uparrow$\\
    \midrule
    \ding{55} & \ding{51} & \ding{55} & 0.847 & 0.911 & 0.852 \\
    \ding{51} & \ding{55} & \ding{55} & 0.888 & 0.936 & 0.890 \\
    \ding{51} & \ding{51} & \ding{55} & \underline{0.891} & \underline{0.938} & \underline{0.893} \\
    \ding{51} & \ding{51} & \ding{51} & \textbf{0.899} & \textbf{0.942} & \textbf{0.901} \\
    \bottomrule
    \end{tabular}
    \label{tab:ablation}
\end{table}

\subsection{Ablation Study}
% Local Geometry-Aware Negative Sampling (LGN)
% Acc.$\downarrow$ & Comp.$\downarrow$ & Overall$\downarrow$

\begin{figure}[tbp]
  \centering
   \includegraphics[width=1\linewidth]{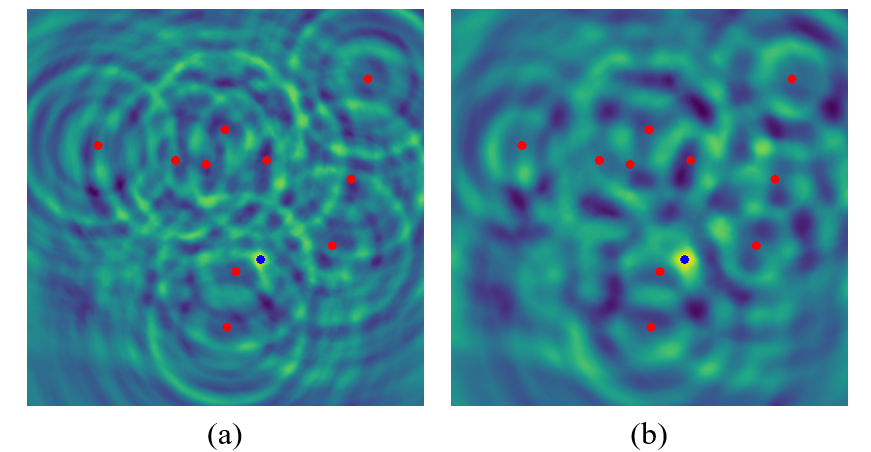}
   \caption{Ablation study on random perturbations. (a) Similarity map generated by SAR2Net trained without random perturbation,  (b) Similarity map by SAR2Net trained with random perturbation, with the query point shown in blue and anchor points in red.} 
   \label{fig:pertu}
\end{figure}

We conduct a comprehensive ablation study to assess the contributions of three key components in SAR2Net: random perturbations (pertu), local geometry-aware negative sampling (LGN), and the second-round refinement (sec).

\begin{figure*}[t]
  \centering
   \includegraphics[width=1\linewidth]{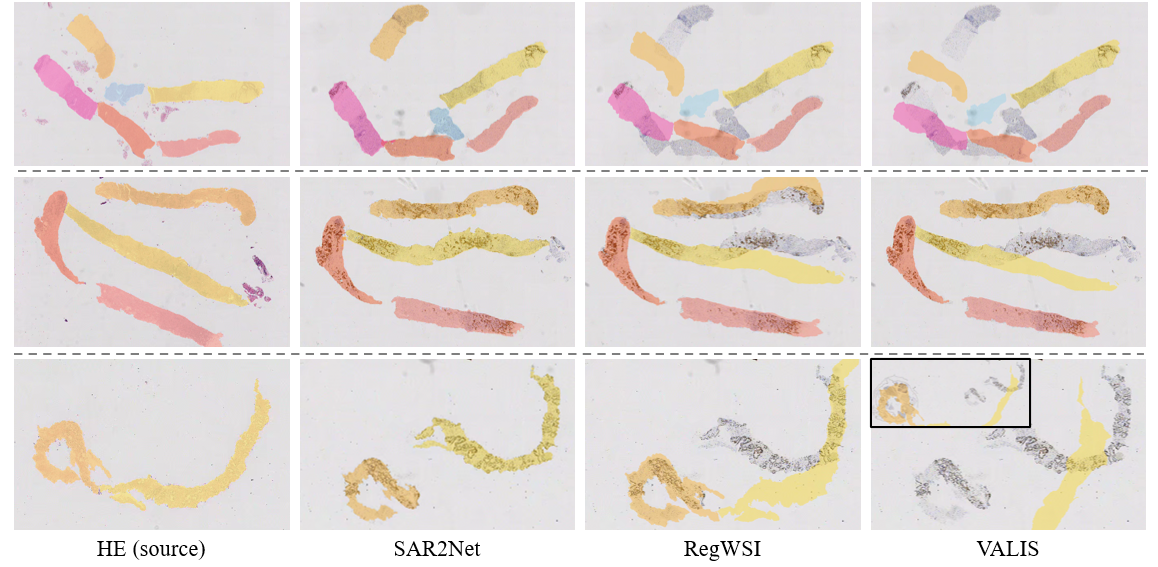}
   \caption{We visualize three representative cases exhibiting fragmented tissues or strong deformations. In the first row, each slide contains multiple tissue fragments that undergo distinct spatial displacements. The second row shows a sample with a tissue piece undergoing large distortion. The third row presents intra-block misalignment, where an additional control tissue on the left (shown in the VALIS result) further distracts the registration. Our method achieves robust spatial correspondence across stains despite these challenges.} 
   \label{fig:result}
\end{figure*}
% 我们可视化了3例存在碎片化组织或严重变形的例子。第一行中的slides由多块组织组成，每块组织在IHC中经历了不同的空间位置变换，第二行中的slides存在一块有着较大扭曲的组织，而第三行中的切片同样存在块内错位，并且在其左侧还有额外的对照组织（在VALIS结果图的左上角有完整的展示，该对照组织影响了VALIS的结果）

\begin{figure}[t]
  \centering
   \includegraphics[width=1\linewidth]{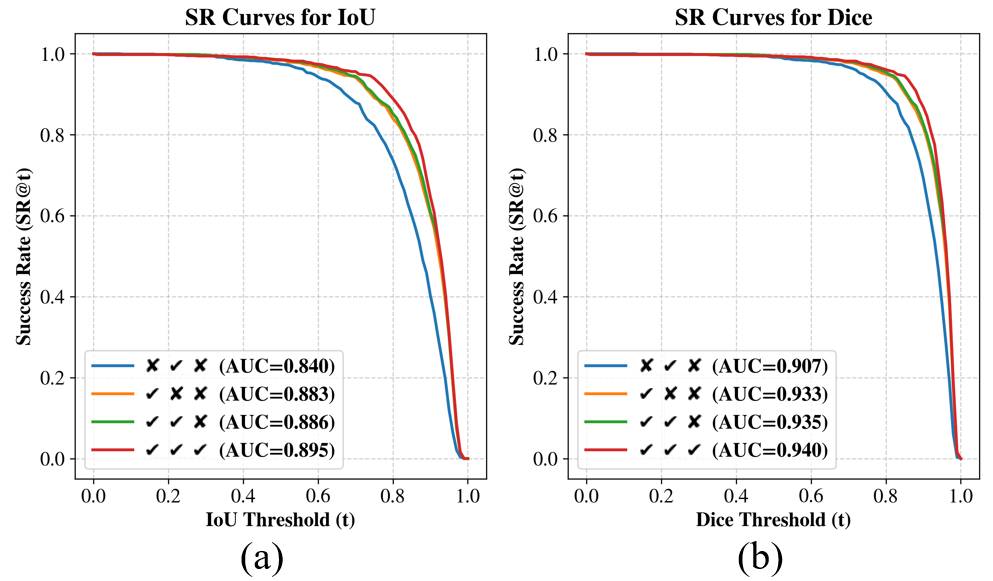}
   \caption{Success rate curves for ablation study. (a) Curves over IoU thresholds. (b) Curves over Dice thresholds. From left to right, each combination of \ding{51} and \ding{55} indicates  whether random perturbations, Local Geometry-Aware Negative Sampling (LGN), and second-round refinement are enabled, respectively. The AUC of the curves are reported on the figures.} 
   \label{fig:sr_ablation}
\end{figure}

As shown in \cref{tab:ablation}, random perturbations provide the most significant performance gains. Incorporating perturbations  during training leads to large improvements in mIoU, mDice, and aw-IoU, indicating substantially enhanced robustness to large and nonlinear deformations. This observation is fully consistent with the success rate curves in \cref{fig:sr_ablation}: without perturbations , the SR curves drop sharply in the mid-to-high threshold ranges (\eg 0.7–0.9), whereas the curves with perturbations remain consistently higher across all thresholds, demonstrating more reliable retrieval of the correct corresponding region.

The similarity maps in \cref{fig:pertu} further illustrate this effect. Without perturbations, the model exhibits clear concentric interference patterns, and the highlighted region collapses into a narrow neighborhood around the query–anchor distance, revealing an undesired tendency toward rigid matching. In contrast, when random perturbations are applied, the model still correctly identifies the corresponding position, but the activation becomes much smoother and more spatially flexible, better reflecting the nonlinear deformations commonly observed in real WSI pairs. This confirms that random perturbations effectively prevent the network from overfitting to rigid transformations and encourage the learning of more realistic geometric variations.

As for the local geometry-aware negative sampling, it plays a more localized and fine-grained role. Adding LGN leads to modest improvements in mIoU and mDice, while at the same time, the SR curves with and without LGN nearly overlap in \cref{fig:sr_ablation}. This indicates that LGN mainly enhances boundary-level geometric discrimination when the query lies close to certain anchors, but has limited influence on the overall large-region alignment.

Finally, incorporating the second-round refinement yields small yet consistent improvements in both average metrics and high-threshold SR values, leading to the best overall performance. This stage refines the residual mis-alignments left by the first round and adjusts boundary details, resulting in more precise final alignment.

\section{Conclusion}
We present SAR2Net, a framework that learns spatially anchored representations for cross-stain region alignment. By modeling the relative geometric relationships between query points and surrounding anchors, SAR2Net enables cross-slide correspondence without any pre-alignment. Through contrastive learning on simulated point coordinates, retrieval-guided region alignment, and multi-level refinement, our method remains robust to severe deformations and fragmented tissue. Experimental results demonstrate that SAR2Net consistently outperforms existing registration frameworks, offering a new perspective for multi-stain integration in computational pathology.

{
    \small
    \bibliographystyle{ieeenat_fullname}
    \bibliography{main}
}


\end{document}